\pdfoutput=1

\documentclass[11pt]{article}

\usepackage[]{naacl2021}

\usepackage{times}
\usepackage{latexsym}

\usepackage[T1]{fontenc}

\usepackage[utf8]{inputenc}

\usepackage{microtype}

\usepackage{graphicx}
\usepackage{amsmath}
\usepackage{booktabs}
\usepackage{array}
\usepackage{amsfonts}
\usepackage{bbm}
\usepackage{enumerate}
\usepackage{multirow}
\usepackage{color}
\usepackage{threeparttable}

\newcommand{\moe}[1]{{\color{black} #1}}

\usepackage[normalem]{ulem}
\usepackage{mathrsfs}
\usepackage{amsthm}
\usepackage{amssymb, bm}
\usepackage{multirow}
\usepackage{booktabs}
\usepackage{enumitem}

\title{
DAGN: Discourse-Aware Graph Network for Logical Reasoning
}

\author{
  Yinya Huang\textsuperscript{1}\thanks{~~This work was done during Yinya Huang's internship in Tencent with L. Wang and M. Fang.} ~~~
  Meng Fang\textsuperscript{2} ~~~
  Yu Cao\textsuperscript{3} ~~~
  Liwei Wang\textsuperscript{4} ~~~
  Xiaodan Liang\textsuperscript{1}\thanks{~~Corresponding Author: Xiaodan Liang.} \\
  
  $^1$Shenzhen Campus of Sun Yat-sen University \\
  $^2$Tencent Robotics X \\
  $^3$School of Computer Science, The University of Sydney \\
  $^4$The Chinese University of Hong Kong \\
  
  \tt yinya.huang@hotmail,
  \tt mfang@tencent.com, \\
  \tt ycao8647@uni.sydney.edu.au, \\
  \tt lwwang@cse.cuhk.edu.hk,
  \tt xdliang328@gmail.com
}

\begin{document}
\maketitle
\begin{abstract}
Recent QA with logical reasoning questions requires passage-level relations among the sentences. However, current approaches still focus on sentence-level relations interacting among tokens. In this work, we explore aggregating passage-level clues for solving logical reasoning QA by using discourse-based information. We propose a discourse-aware graph network (DAGN) that reasons relying on the discourse structure of the texts. The model encodes discourse information as a graph with elementary discourse units (EDUs) and discourse relations, and learns the discourse-aware features via a graph network for downstream QA tasks. Experiments are conducted on two logical reasoning QA datasets, ReClor and LogiQA, and our proposed DAGN achieves competitive results. The source code is available at \href{https://github.com/Eleanor-H/DAGN}{https://github.com/Eleanor-H/DAGN}.

\end{abstract}

\section{Introduction}
\vspace{-1mm}
A variety of QA datasets have promoted the development of reading comprehensions, for instance, SQuAD \cite{rajpurkar2016squad}, HotpotQA \cite{yang2018hotpotqa}, DROP \cite{dua2019drop}, \moe{and so on}. 
Recently, QA datasets with more complicated reasoning types, i.e., logical reasoning, are also introduced, such as ReClor \cite{yu2020reclor} and LogiQA \cite{liu2020logiqa}. 
The logical questions are taken from standardized exams such as GMAT and LSAT, and require QA models to read complicated argument passages and 
\moe{identify logical relationships therein. For example, selecting a correct assumption that supports an argument, or finding out a claim that weakens an argument in a passage.} 
Such logical reasoning is beyond the capability of most of the previous QA models which focus on reasoning with entities or numerical keywords.

A main challenge for the QA models is to uncover the logical structures under passages, such as identifying claims or hypotheses, or pointing out flaws in arguments. To achieve this, the QA models should first be aware of logical units, which can be sentences or clauses or other meaningful text spans, then identify the logical relationships between the units. 
However, the logical structures are usually hidden \moe{and difficult to be extracted}, \moe{and most datasets do not provide such logical structure annotations.}

An intuitive idea for unwrapping such logical information is \moe{using} discourse relations. For instance, 
\moe{as a conjunction, ``because'' indicates a causal relationship, whereas ``if'' indicates a hypothetical relationship.} However, such discourse-based information is seldom considered in logical reasoning tasks. 
\moe{Modeling logical structures is still lacking in logical reasoning tasks}, while current opened methods use contextual pre-trained models \cite{yu2020reclor}. 
\moe{Besides, previous graph-based methods \cite{ran2019numnet, chen2020question} that construct entity-based graphs are not suitable for logical reasoning tasks because of different reasoning units.}



\begin{figure*}[t!]
    \setlength{\belowcaptionskip}{-0.35cm}
    \centering
    \includegraphics[width=\textwidth]{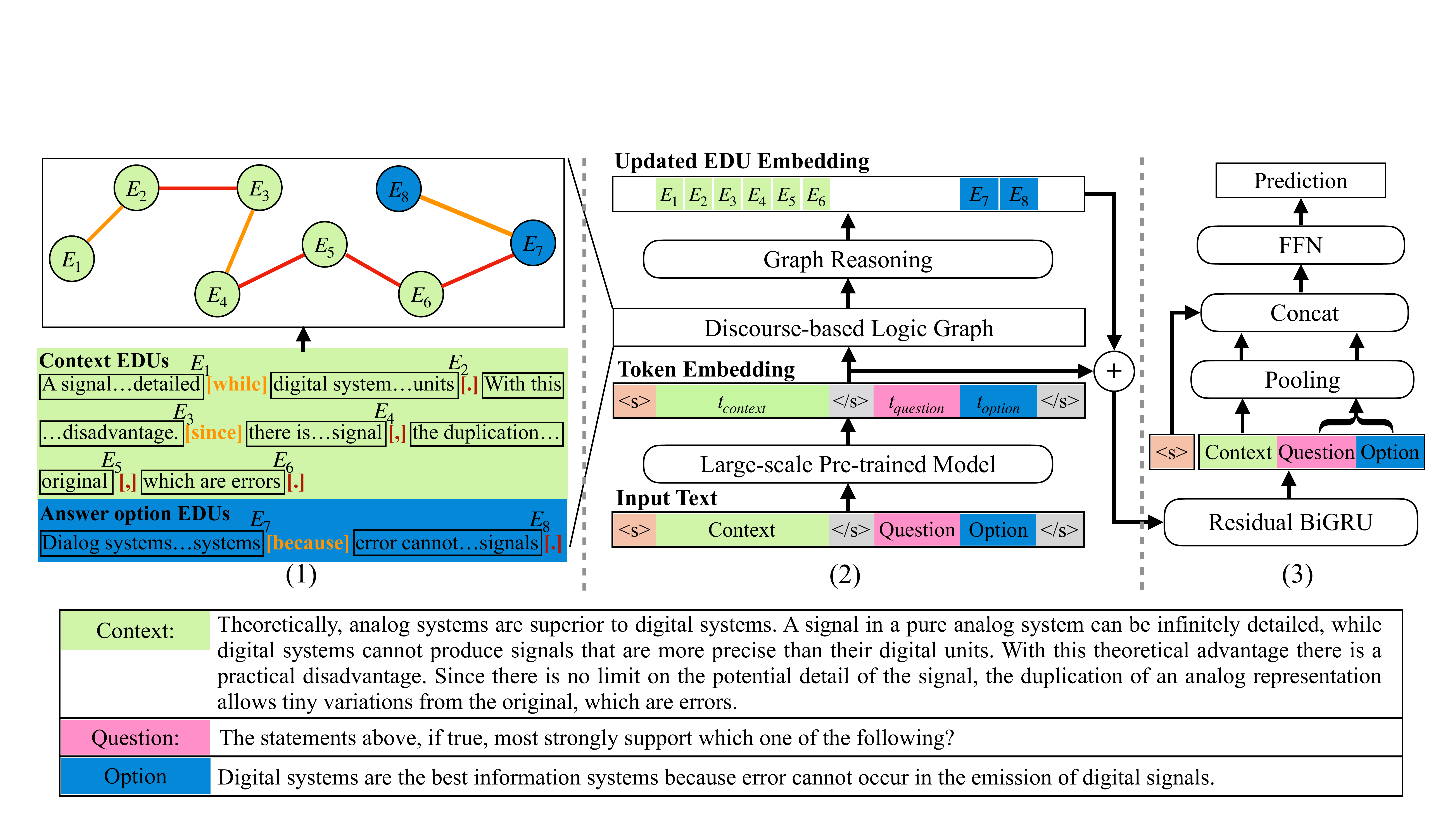}
    \caption{
    The architecture of our proposed method with an example below. 
    }
    \label{fig:model}
\end{figure*}

In this paper, we propose a new approach to solve logical reasoning QA tasks by incorporating 
discourse-based information.
First, \moe{we construct discourse structures}. We use discourse relations from the Penn Discourse TreeBank 2.0 (PDTB 2.0) \cite{prasad2008penn} as delimiters to split texts into elementary discourse units (EDUs). A logic graph is constructed in which EDUs are nodes and discourse relations are edges.
Then, we propose a Discourse-Aware Graph Network (DAGN) for learning high-level discourse features \moe{to represent passages.}
The discourse features are incorporated with the contextual token features from pre-trained language models.
With the enhanced features, DAGN predicts answers to logical questions.
Our experiments show that DAGN surpasses current opened methods
on two \moe{recent} logical reasoning QA datasets, ReClor and LogiQA.

Our main contributions are three-fold:
\begin{itemize}
\item We propose to construct logic graphs from texts by using discourse relations as edges and elementary discourse units as nodes. 
\item We obtain discourse features via graph neural networks to facilitate logical reasoning in QA models. 
\item We show the effectiveness of using logic graph and feature enhancement by noticeable improvements on two datasets, ReClor and LogiQA.
\end{itemize}

\section{Method}
\vspace{-1mm}


Our intuition is to explicitly use discourse-based information to mimic the human reasoning process for logical reasoning questions.
The questions are in multiple choices format, which means given a triplet (context, question, answer options), models answer the question by selecting the correct answer option. 
Our framework is shown in Figure~\ref{fig:model}.
We first construct a discourse-based logic graph from the raw text. 
Then we conduct reasoning via graph networks to learn and update the discourse-based features, which are incorporated with the contextual token embeddings for downstream answer prediction.


\subsection{Graph Construction}
\moe{Our} discourse-based logic graph is constructed via two steps: delimiting text into elementary discourse units (EDUs) and forming the graph using their relations as edges, as illustrated in Figure~\ref{fig:model}(1).

\paragraph{Discourse Units Delimitation} 

It is studied that clause-like text spans delimited by discourse relations can be discourse units that reveal the rhetorical structure of texts \cite{mann1988rhetorical,prasad2008penn}.
We further observe that such discourse units are essential units in logical reasoning, such as being assumptions or opinions.
As the example shown in Figure~\ref{fig:model}, the ``\textit{while}'' in the context indicates a comparison between the attributes of ``\textit{pure analog system}'' and that of ``\textit{digital systems}''.
The ``\textit{because}'' in the option provides evidence ``\textit{error cannot occur in the emission of digital signals}'' to the claim ``\textit{digital systems are the best information systems}''.

\moe{We use PDTB 2.0~\cite{prasad2008penn} to help drawing discourse relations.}
PDTB 2.0 contains discourse relations that are manually annotated on the 1 million Wall Street Journal (WSJ) corpus and are broadly characterized into ``Explicit'' and ``Implicit'' connectives.
The former apparently presents in sentences such as discourse adverbial ``\textit{instead}'' or subordinating conjunction ``\textit{because}'',
whereas the latter are inferred by annotators between successive pairs of text spans split by punctuation marks such as ``.'' or ``;''.
We simply take all the ``Explicit'' connectives as well as common punctuation marks to form our discourse delimiter library (details are given in Appendix~\ref{sec:library}), with which we delimit the texts into EDUs.
For each data sample, we segment the context and options, ignoring the question since the question usually does not carry logical content.

\paragraph{Discourse Graph Construction} 
We define the discourse-based graphs with EDUs as nodes, the ``Explicit'' connectives as well as the punctuation marks as two types of edges.
We assume that each connective or punctuation mark connects the EDUs before and after it.
For example, the option sentence in Figure~\ref{fig:model}
is delimited into two EDUs, $\text{EDU}_7=$``\textit{digital systems are the best information systems}'' and $\text{EDU}_8=$``\textit{error cannot occur in the emission of digital signals}'' by the connective $r=$``\textit{because}''. Then the returned triplets are $(\text{EDU}_7, r, \text{EDU}_8)$ and $(\text{EDU}_8, r, \text{EDU}_7)$.
%
For each data sample with the context and multiple answer options, we separately construct graphs corresponding to each option, with EDUs in the same context and every single option. The graph for the single option $k$ is denoted by $\mathcal{G}^k = (\mathcal{V}^k, \mathcal{E}^k)$.

\subsection{Discourse-Aware Graph Network}
We present the Discourse-Aware Graph Network (DAGN) that uses the constructed graph to exploit discourse-based information for answering logical questions.
\moe{
It consists of three main components: an EDU encoding module, a graph reasoning module, and an answer prediction module.}
The former two are demonstrated in Figure~\ref{fig:model}(2), whereas the final component is in Figure~\ref{fig:model}(3).




\paragraph{EDU Encoding}
An EDU span embedding is obtained from its token embeddings.
\moe{
There are two steps. 
First, similar to previous works \cite{yu2020reclor,liu2020logiqa},} 
we encode such input sequence ``\texttt{<s> context </s> question || option </s>}'' into contextual token embeddings with pre-trained language models,
where \texttt{<s>} and \texttt{</s>} are the special tokens for RoBERTa \cite{liu2019roberta} model, and \texttt{||} denotes concatenation. \moe{Second,} given the token embedding sequence $\{\mathbf{t}_1, \mathbf{t}_2,...,\mathbf{t}_L\}$, the $n$-th EDU embedding is obtained by $\mathbf{e}_n=\sum\limits_{l \in {S_n}} {\mathbf{t}_l}$, where $S_n$ is the set of token indices belonging to $n$-th EDU.


\paragraph{\moe{Graph Reasoning}} \moe{After EDU encoding, DAGN performs reasoning over the discourse graph. 
Inspired by previous graph-based models~\cite{ran2019numnet,chen2020question},
we also learn graph node representations to obtain higher-level features. However, we consider different graph construction and encoding.
Specifically, let $\mathcal{G}^k = (\mathcal{V}^k, \mathcal{E}^k)$ denote a graph corresponding to the $k$-th option in answer choices.
For each node $v_i \in \mathbf{V}$, the node embedding $\mathbf{v}_i$ is initialized with the corresponding EDU embedding $\textbf{e}_i$. 
$\mathcal{N}_i = \{j|(v_j, v_i)\in\mathcal{E}^k \}$ indicates the neighbors of node $v_i$.
$\mathbf{W}^{r_{ji}}$ is the adjacency matrix for one of the two edge types, 
where $r_E$ indicates graph edges corresponding to the explicit connectives, 
and $r_I$ indicates graph edges corresponding to punctuation marks.}

The model first calculates weight $\alpha_i$ for each node with a linear transformation and a sigmoid function $\alpha_i=\sigma(\mathbf{W}^{\alpha}(\mathbf{v}_i)+b^{\alpha})$, then conducts message propagation with the weights:
\begin{equation}
    \tilde{\mathbf{v}}_i = \frac{1}{|\mathcal{N}_i|}(\sum_{j\in\mathcal{N}_i}\alpha_j\mathbf{W}^{r_{ji}}\mathbf{v}_j),  r_{ji} \in \{r_E, r_I\}
\end{equation}
\noindent where $\tilde{\mathbf{v}}_i$ is the message representation of node $v_i$. 
$\alpha_j$ and $\mathbf{v}_j$ are the weight and the node embedding of $v_j$ respectively.

After the message propagation, the node representations are updated with the initial node embeddings and the message representations by
\begin{equation}
    \mathbf{v}^{\prime}_i = \text{ReLU}(\mathbf{W}^{u}\mathbf{v}_i + \tilde{\mathbf{v}}_i + \mathbf{b}^{u}),
\end{equation}
where $\mathbf{W}^{u}$ and $\mathbf{b}^{u}$ are weight and bias respectively.
The updated node representations $\mathbf{v}^{\prime}_i$ will be used to enhance the contextual token embedding via summation in corresponding positions. Thus $\mathbf{t}_l^{\prime} = \mathbf{t}_l + \mathbf{v}^{\prime}_n$, where $l \in S_n$ and $S_n$ is the corresponding token indices set for $n$-th EDU.

\paragraph{Answer Prediction}
\moe{
The probabilities of options are obtained by feeding the discourse-enhanced token embeddings into the answer prediction module.
The model is end-to-end trained using cross entropy loss.}
Specifically, the embedding sequence first goes through a layer normalization~\cite{ba2016layer}, then a bidirectional GRU~\cite{cho2014properties}. The output embeddings are then added to the input ones as the residual structure~\cite{he2016deep}. 
We finally obtain the encoded sequence after another layer normalization on the added embeddings.

We then merge the high-level discourse features and the low-level token features.
Specifically, the variant-length encoded context sequence, question-and-option sequence are pooled via weighted summation wherein the weights are softmax results of a linear transformation of the sequence,
resulting in single feature vectors separately. 
We concatenate them with ``\texttt{<s>}'' embedding from the backbone pre-trained model, and feed the new vector into a two-layer perceptron with a GELU activation~\cite{hendrycks2016gaussian}
to get the output features for classification.







\section{Experiments}
\vspace{-1mm}
We evaluate the performance of DAGN on two logical reasoning datasets, ReClor~\cite{yu2020reclor} and LogiQA~\cite{liu2020logiqa}, and conduct ablation study on graph construction and graph network. The implementation details are shown in Appendix~\ref{sec:appendix_implementation_details}.

\subsection{Datasets}
\vspace{-1mm}
ReClor contains 6,138 questions modified from standardized tests such as GMAT and LSAT, which are split into train / dev / test sets with 4,638 / 500 / 1,000 samples respectively. The training set and the development set are available.
The test set is blind and hold-out, and split into an EASY subset and a HARD subset according to the performance of BERT-base model~\cite{devlin2019bert}. 
The test results are obtained by submitting the test predictions to the leaderboard.
%
LogiQA consists of 8,678 questions that are collected from National Civil Servants Examinations of China and manually translated into English by professionals. 
The dataset is randomly split into train / dev / test sets with 7,376 / 651 / 651 samples respectively.
Both datasets contain multiple logical reasoning types. 

\begin{table}
    \setlength{\belowcaptionskip}{-0.4cm}
    \footnotesize
    \centering
    \begin{threeparttable}
    \begin{tabular}{
    p{0.135\textwidth}
    p{0.050\textwidth}<\centering
    p{0.050\textwidth}<\centering
    p{0.055\textwidth}<\centering
    p{0.058\textwidth}<\centering
    }
    \toprule
     \textbf{Methods} & \textbf{Dev} & \textbf{Test} & \textbf{Test-E} & \textbf{Test-H} \\
     \midrule
     BERT-Large & 53.80 & 49.80 & 72.00 & 32.30 \\
     XLNet-Large & 62.00 & 56.00 & 75.70 & 40.50 \\
     RoBERTa-Large & 62.60 & 55.60 & 75.50 & 40.00 \\
     DAGN & \textbf{65.20} & \textbf{58.20} & \textbf{76.14} & \textbf{44.11} \\
     DAGN (Aug) & \textbf{65.80} & \textbf{58.30} & \textbf{75.91} & \textbf{44.46} \\
     \bottomrule
    \end{tabular}
    \begin{tablenotes}
        \footnotesize
        \item[*] The results are taken from the ReClor paper. 
        \item[*] DAGN ranks the 1st on the public ReClor leaderboard\footnotemark until 17th Nov., 2020 before submitting it to NAACL. Until now, we find that several better results appeared in the leaderboard and they are not opened. 
      \end{tablenotes}
  \end{threeparttable}
    \caption{Experimental results (accuracy \%) of DAGN compared with baseline models on ReClor dataset.
    Test-E = Test-EASY, Test-H = Test-HARD.}
    \label{tab:res_reclor}
\end{table}

\begin{table}
    \setlength{\belowcaptionskip}{-0.1cm}
    \footnotesize
    \centering
    \begin{tabular}{
    p{0.15\textwidth}
    p{0.057\textwidth}<\centering
    p{0.057\textwidth}<\centering
    }
    \toprule
     \textbf{Methods} & \textbf{Dev} & \textbf{Test} \\
     \midrule
     BERT-Large & 34.10	& 31.03 \\
     RoBERTa-Large & 35.02 & 35.33 \\
     DAGN & \textbf{35.48} & \textbf{38.71} \\
     DAGN (Aug) & \textbf{36.87} & \textbf{39.32} \\
    \bottomrule
    \end{tabular}
    \caption{Experimental results (accuracy \%) of DAGN compared with baseline models on LogiQA dataset.}
    \label{tab:res_logiqa}
\end{table}

\subsection{Results}
\vspace{-1mm}
The experimental results are shown in Tables~\ref{tab:res_reclor} and~\ref{tab:res_logiqa}. 
Since there is no public method for both datasets, we compare DAGN with the baseline models.
As for DAGN, we fine-tune RoBERTa-Large as the backbone. 
DAGN (Aug) is a variant that augments the graph features.

DAGN reaches 58.20\% of test accuracy on ReClor.
DAGN (Aug) reaches 58.30\%, therein 75.91\% on EASY subset, and 44.46\% on HARD subset. Compared with RoBERTa-Large, the improvement on the HARD subset is remarkably 4.46\%. This indicates that the incorporated discourse-based information supplements the \moe{shortcoming of the baseline model}, and that the discourse features are beneficial for such logical reasoning.
Besides, DAGN and DAGN (Aug) also outperform the baseline models on LogiQA, especially showing 4.01\% improvement over RoBERTa-Large on the test set.

\begin{table}[t]
    \setlength{\belowcaptionskip}{-0.4cm}
    \footnotesize
    \centering
    \begin{tabular}{
    p{0.36\textwidth}
    p{0.05\textwidth}<\centering
    }
    \toprule
     \textbf{Methods} & \textbf{Dev}  \\
     \midrule
     DAGN & \textbf{65.20} \\
     \textbf{\textit{ablation on nodes}} \\
     DAGN - clause nodes & 64.40 \\
     DAGN - sentence nodes & 64.40 \\
     \textbf{\textit{ablation on edges}} \\
     DAGN - single edge type & 64.80 \\
     DAGN - fully connected edges & 61.60 \\
     \textbf{\textit{ablation on graph reasoning}} \\
     DAGN w/o graph module & 64.00 \\
     \bottomrule
    \end{tabular}
    \caption{Ablation study results (accurcy \%) on ReClor development set.}
    \label{tab:ablation}
\end{table}

\subsection{Ablation Study}
\vspace{-1mm}
We conduct ablation study on graph construction details as well as the graph reasoning module.
The results are reported in Table~\ref{tab:ablation}.

\footnotetext{\url{https://bit.ly/2UOQfaS}}

\paragraph{Varied Graph Nodes}
We first use clauses or sentences in substitution for EDUs as graph nodes. 
For clause nodes, we simply remove ``Explicit'' connectives during discourse unit delimitation. So that the texts are just delimited by punctuation marks.
For sentence nodes, we further reduce the delimiter library to solely period (``.''). 
Using the modified graphs with clause nodes or coarser sentence nodes, the accuracy of DAGN drops to 64.40\%.
This indicates that clause or sentence nodes carry less discourse information and act poorly as logical reasoning units.
%
%

\paragraph{Varied Graph Edges} 
We make two changes of the edges: (1) modifying the edge type, (2) modifying the edge linking. 
For edge type, all edges are regarded as a single type.
For edge linking, we ignore discourse relations and connect every pair of nodes, turning the graph into fully-connected.
The resulting accuracies drop to 64.80\% and 61.60\% respectively.
It is proved that in the graph we built, edges link EDUs in reasonable manners, which properly indicates the logical relations.

\paragraph{Ablation on Graph Reasoning}
\moe{We remove the graph module from DAGN and give a comparison.}
This model solely contains an extra prediction module than the baseline. The performance on ReClor dev set is between the baseline model and DAGN. Therefore, despite the prediction module benefits the accuracy, the lack of graph reasoning leads to the absence of discourse features and degenerates the performance. It demonstrates the necessity of discourse-based structure in logical reasoning.

\section{Related Works}
\vspace{-1mm}
Recent datasets for reading comprehension tend to be more complicated and require models' capability of reasoning. For instance, HotpotQA \cite{yang2018hotpotqa}, WikiHop \cite{welbl2018constructing}, OpenBookQA \cite{mihaylov2018can}, and MultiRC \cite{khashabi2018looking} require the models to have multi-hop reasoning.
DROP \cite{dua2019drop} and MA-TACO \cite{zhou-etal-2019-going} need the models to have numerical reasoning.
WIQA \cite{tandon2019wiqa} and CosmosQA \cite{huang2019cosmos} require causal reasoning that the 
models can understand the counterfactual hypothesis or find out the cause-effect relationships in events. 
However, the logical reasoning datasets \cite{yu2020reclor, liu2020logiqa} 
require the models to have the logical reasoning capability of uncovering the inner logic of texts. 

Deep neural networks are used for reasoning-driven RC.
Evidence-based methods \cite{madaan2020eigen, huang2020rem, rajagopal2020if} generate explainable evidence from a given context as the backup of reasoning. 
%
Graph-based methods \cite{qiu-etal-2019-dynamically,de2019question,cao2019bag, ran2019numnet, chen-etal-2020-question,yunqiu2020deep, zhang2020graph} explicitly model the reasoning process with constructed graphs, then learn and update features through message passing based on graphs. 
There are also other methods such as neuro-symbolic models \cite{saha2021weakly} and adversarial training \cite{pereira-etal-2020-adversarial}. 
%
Our paper uses a graph-based model. However, for uncovering logical relations, graph nodes and edges are customized with discourse information. 

Discourse information provides a high-level understanding of texts and hence is beneficial for many of the natural language tasks, for instance, text summarization \cite{cohan-etal-2018-discourse, joty-etal-2019-discourse, xu-etal-2020-discourse, feng2020dialogue}, neural machine translation \cite{voita2018context}, and coherent text generation \cite{wang2020consistency, bosselut2018discourse}. 
There are also discourse-based applications for reading comprehension. DISCERN \cite{gao2020discern} segments texts into EDUs and learns interactive EDU features.
\citet{mihaylov-frank-2019-discourse} provide additional discourse-based annotations and encodes them with discourse-aware self-attention models. 
%
Unlike previous works, DAGN first uses discourse relations as graph edges connecting EDUs for texts, then learns the discourse features via message passing with graph neural networks. 

\section{Conclusion}
\vspace{-1mm}
\moe{In this paper, we introduce a Discourse-Aware Graph Network (DAGN) to addressing logical reasoning QA tasks. We first treat elementary discourse units (EDUs) that are split by discourse relations as basic reasoning units. We then build discourse-based logic graphs with EDUs as nodes and discourse relations as edges. DAGN then learns the discourse-based features and enhances them with contextual token embeddings. DAGN reaches competitive performances on two recent logical reasoning datasets ReClor and LogiQA.}

\section*{Acknowledgements}
\vspace{-1mm}
The authors would like to thank Wenge Liu, Jianheng Tang, Guanlin Li and Wei Wang for their support and useful discussions. 
This work was supported in part by National Natural Science Foundation of China (NSFC) under Grant No.U19A2073 and No.61976233, Guangdong Province Basic and Applied Basic Research (Regional Joint Fund-Key) Grant No.2019B1515120039,  Shenzhen Basic Research Project (Project No. JCYJ20190807154211365), Zhijiang Lab’s Open Fund (No. 2020AA3AB14) and CSIG Young Fellow Support Fund.


\bibliography{custom}
\bibliographystyle{acl_natbib}

\newpage
\appendix
\section{Discourse Delimiter Library}
\label{sec:library}
Our discourse delimiter library consists of two parts, the ``Explicit'' connectives annotated in Penn Discourse TreeBank 2.0 (DPTB 2.0)~\cite{prasad2008penn}, as well as a set of punctuation marks. The overall discourse delimiters used in our method are presented in Table~\ref{tab:delimeter}.

\begin{table}[bh!]
    \footnotesize
    \setlength{\belowcaptionskip}{-0.15cm}
	\begin{tabular}{
    |p{0.45\textwidth}<\centering|
    }
    \hline
    \textbf{Explicit Connectives} \\
    \hline
    'once', 'although', 'though', 'but', 'because', 'nevertheless', 'before', 'for example', 'until', 'if', 'previously', 'when', 'and', 'so', 'then', 'while', 'as long as', 'however', 'also', 'after', 'separately', 'still', 'so that', 'or', 'moreover', 'in addition', 'instead', 'on the other hand', 'as', 'for instance', 'nonetheless', 'unless', 'meanwhile', 'yet', 'since', 'rather', 'in fact', 'indeed', 'later', 'ultimately', 'as a result', 'either or', 'therefore', 'in turn', 'thus', 'in particular', 'further', 'afterward', 'next', 'similarly', 'besides', 'if and when', 'nor', 'alternatively', 'whereas', 'overall', 'by comparison', 'till', 'in contrast', 'finally', 'otherwise', 'as if', 'thereby', 'now that', 'before and after', 'additionally', 'meantime', 'by contrast', 'if then', 'likewise', 'in the end', 'regardless', 'thereafter', 'earlier', 'in other words', 'as soon as', 'except', 'in short', 'neither nor', 'furthermore', 'lest', 'as though', 'specifically', 'conversely', 'consequently', 'as well', 'much as', 'plus', 'and', 'hence', 'by then', 'accordingly', 'on the contrary', 'simultaneously', 'for', 'in sum', 'when and if', 'insofar as', 'else', 'as an alternative', 'on the one hand on the other hand' \\
    \hline\hline
    \textbf{Punctuation Marks} \\
    \hline
    '.', ',', ';', ':' \\
    \hline
    \end{tabular}
    \caption{The discourse delimiter library in our implementation.}
    \label{tab:delimeter}
\end{table}



\section{Implementation Details}
\label{sec:appendix_implementation_details}
We fine-tune RoBERTa-Large~\cite{liu2019roberta} as the backbone pre-trained language model for DGAN, which contains 24 hidden layers with hidden size 1024. 
The overall model is end-to-end trained and updated by Adam \cite{kingma2014adam} optimizer with an overall learning rate of 5e-6 and a weight decay of 0.01. The overall dropout rate is 0.1. The maximum sequence length is 256. 
We tune the model on the dev set to obtain the best iteration steps of graph reasoning, which is 2 for ReClor data, and 3 for LogiQA data.
The model is trained for 10 epochs with a batch size of 16 on Nvidia Tesla V100 GPU.

For the answer prediction module,  the hidden size of GRU is the same as the token embeddings in the pre-trained language model, which is 1024.
The two-layer perceptron first projects the concatenated vectors with a hidden size of 1024 $\times$ 3 to 1024, then project 1024 to 1.

\end{document}